\documentclass[twoside,11pt]{article}

\usepackage{jmlr2e}

\usepackage[utf8]{inputenc} 
\usepackage[T1]{fontenc}    
\usepackage{hyperref}       
\usepackage{url}            
\usepackage{booktabs}       
\usepackage{amsfonts}       
\usepackage{nicefrac}       
\usepackage{microtype}      
\usepackage{graphicx}
\usepackage{caption}
\usepackage{subcaption}
\usepackage{gensymb}
\usepackage[parfill]{parskip}

\usepackage{hyperref}

\usepackage{multirow}

\usepackage{algorithm}
\usepackage[noend]{algpseudocode}

\ShortHeadings{Graphical RNN Models}{Graphical RNN models}
\firstpageno{1}

\begin{document}

\title{Graphical RNN Models}

\author{Ashish Bora \\
  Department of Computer Science\\
  University of Texas at Austin\\
  \texttt{ashish.bora@utexas.edu} \\
  \AND
  Sugato Basu \\
  Google \\
  \texttt{sugato@google.com} \\
  \AND
  Joydeep Ghosh \\
  Department of Electrical and Computer Engineering \\
  University of Texas at Austin\\
  \texttt{jghosh@utexas.edu} \\
}

\editor{}

\maketitle

\begin{abstract}
Many time series are generated by a set of entities that interact with one another over time. This paper introduces a broad, flexible framework to learn from multiple inter-dependent time series generated by such entities.  Our framework explicitly models the entities and their interactions through time. It achieves this by building on the capabilities of Recurrent Neural Networks, while also offering several ways to incorporate domain knowledge/constraints into the model architecture. The capabilities of our approach are showcased through an application to weather prediction, which shows gains over strong baselines.
\end{abstract}


\section{Introduction} \label{Intro}
Time series data in many domains have an underlying relational structure in addition to the inherent temporal structure. For example, when we model the behavior of users in a social network, the browsing or click patterns form the time series while the relation between users (e.g., likes, friend links, participating in same forums) provides the underlying relational structure. Similarly, if we consider a wireless sensor recording weather-related information (e.g., temperature or rainfall levels), then the underlying physical layout of the wireless sensor network provides the relational structure while the data stream from each sensor gives the time series. Robotics is another domain where several agents (e.g., individual robots) can interact with each other for a particular task (e.g., in a group search-and-rescue mission).

In each of these cases, individual time series data is generated by an entity in an environment, while there are relationships between multiple entities in the same environment. These entities interact each other over a period of time, and thus the time series generated by these entities are correlated and influenced by relationships and interactions with other entities. The interactions between entities provides vital information about the joint evolution of the time series data, which is not explicitly available if we consider each entity separately. This information can thus help us better model the dynamics of such systems and is thus potentially useful for learning tasks that use the time series data. The system as a whole evolves jointly through time, so if we are able to learn the dynamics of the underlying process generating the observations (e.g., model the fact that the temporal measurements are spatially correlated), then it could be helpful in making better future predictions.

For modeling and learning using data with such properties, we start by defining a scenario: a sequential interaction of a set of entities with each other over time. We observe a scenario by gathering a set of measurements for each entity, at each time step. Since the entities interact with each other, the measurements are influenced not only by the state of the corresponding entity, but also other entities that it interacts with. The sequence of such observations over time consists of data for a particular scenario. We assume that our dataset consists of multiple such scenarios, each possibly with a different set of entities and relationships between them. In this paper, our goal is to build a model that can learn the dynamics of the network of interactions in such datasets so that it generalizes to novel scenarios.

Recurrent Neural Networks (RNNs) are powerful models for learning from sequential data. They have been successfully applied to a wide range of challenging sequential tasks like character prediction (\cite{sutskever2011generating}), machine translation (\cite{blunsom2013recurrent}), handwriting generation (\cite{graves2013generating}) and autoregressive image modeling (\cite{van2016pixel}). Despite their capabilities, RNNs cannot directly be used in our setting since (a) there is no obvious way to incorporate the graphical structure of the data into a RNN, (b) the number of entities may change between scenarios, thus changing the input dimension per time step, and (c) The interactions may change between scenarios. With this motivation, we propose a new framework to design recurrent model architectures that we call Graphical RNN models (gRNN for short). In addition to addressing the limitations of RNNs described above, our framework exhibits a flexible mechanism of incorporating domain knowledge into the network, similar to a graphical model.

The rest of the paper is organized as follows: In Section~\ref{gRNNFramework}, we describe our framework and some of its salient features. In Section~\ref{Formal}, we formally define the model and outline the inference algorithm, i.e.\ the forward pass. In Section~\ref{RelWork}, we present a brief survey of related work and highlight some of the differences of existing approaches as compared to our framework. In Section~\ref{Synth}, we present experiments on a synthetic dataset with toy models designed to test the basic learning ability of models created using our framework. In Section~\ref{Weather}, we show that on a weather prediction task, models built using our framework achieve superior performance as compared to several standard approaches. Finally, we discuss a few promising directions of future research in Section~\ref{DiscussFuture} and conclude in the final section.

\section{gRNN framework}
\label{gRNNFramework}

In our setting, data in every scenario consists of a set of entities interacting with each other and producing a sequence of observations. To mathematically represent this process within a scenario, we construct a graph where each node represents an entity and edges denote relationships between them. We note that the size and structure of this graph is fixed within a given scenario but may change across scenarios. Thus, generalization to novel scenarios can equivalently be seen as generalization to new graphs. We call this spatial generalization. Additionally, we would like our models to learn the dynamics across time within a scenario. We call this \emph{temporal generalization}. Thus our goal is to develop a flexible framework that affords a natural way to design models capable of \emph{spatio-temporal generalization}.

Our framework thus builds on the powerful temporal generalization capabilities of RNNs, while augmenting them to be able to perform spatial generalization as well. We call our framework graphical RNN models (gRNN) which is described in the next few sections.

\subsection{Core structure}

For a given scenario, in the interaction graph as described above, we put a RNN at each node. At every time step, the RNNs take as input (1) the features associated with the entity representing the node, (2) the observation vector at that node at the current time step, (3) its own hidden state at previous time step, and (4) multiple ``summaries of the activity near it'' from previous time step (we will shortly discuss how summaries are computed).

Based on RNN outputs we get predictions at each node, at each time step. Using these predictions and the ground truth targets we compute the loss per node, per time step and aggregating them gives us the overall loss. By unrolling through time, the aggregated loss can be computed using the forward pass (Algorithm~\ref{gRNN_alg}). The derivatives of the aggregate loss are then back-propagated to get gradients with respect to all model parameters. These gradients can be used with any general purpose gradient based learning algorithm like SGD or Adagrad (\cite{duchi2011adaptive}) to learn all model parameters jointly. We note that our framework can work with any type of recurrent network module.

\subsection{Summary computation} \label{summary-desc}

Our framework uses the graphical structure through summary computation. The purpose of summary input to RNN is to provide a compact representation of the context in which the corresponding entity operates. Since the interactions are captured by the neighborhood of the node in the graph, our context depends on this neighborhood. For popular RNN architectures, such as LSTM (\cite{hochreiter1997long}), the hidden state of stores useful information about the history of inputs it has seen so far. Thus, for each node, the summary is computed as a function of the hidden states of RNNs on the nodes adjacent to it in the graph. We shall use the hidden states from the previous time step to compute the summary for the next time step.

The summary should be neighborhood order agnostic. Thus, we consider only those transformations that are set functions of hidden state vectors. Set functions are naturally permutation invariant and thus satisfy our criteria of being neighborhood order agnostic. It is also important to have the empty set in the domain of the set function to be able to handle nodes that are not connected to others.

The summary should also have fixed dimension because RNNs at each node requires fixed input dimension per time step. Thus, we consider only those set functions that map their input set to a vector with fixed output dimension. This constraint ensures that irrespective of the incoming degree of the node, the output has a fixed dimension.

In this paper, we restrict all RNNs to have the same dimension for the hidden state. Although not strictly necessary, it simplifies the formulation. With this assumption, addition, point-wise multiplication, point-wise maximum are some examples of simple set functions that satisfy our requirements.

Thus, for a summary computation we need a graph and an associated transformation which satisfies the properties outlined above. At every time step, for every node in the graph, we construct a set consisting of hidden state vectors of its neighbors (as given by the graph). The summary is then the output of the transformation applied to this set.

\subsection{Incorporating Domain Knowledge}
One of the main highlights of our framework is that it allows injection of domain knowledge much like a graphical model, but also uses RNNs to learn the temporal dynamics. This makes the gRNN framework quite general and flexible, yet very powerful. The injection of domain knowledge is done through the following key ideas:
\begin{enumerate}
    \item {
        \textbf{Equivalence classes}: Often, entities in a scenario will have similar roles and thus expected to have similar dynamics. To incorporate this knowledge, we partition all nodes into equivalence classes and share the RNN parameters within the class. The equivalence class mapping is assumed to be model input. Parameter sharing also improves the statistical efficiency of learning and helps avoid overfitting. Using equivalence classes also suggests a natural way to generalize to novel scenarios -- for every new node in the novel scenario, we reuse the parameters from the equivalence class for the entity at that node.
    }
    \item {
        \textbf{Multiple summaries to model different relationships}: Interactions between entities can be of several types. To represent different kinds of relationships, instead of using a single graph, we propose to use multiple underlying graphs, one for each relationship type. Every relationship type also comes with a corresponding set function used to compute the summaries. An order among the summary types is predefined and all summaries are concatenated in that order at every node to form the complete summary which is then fed into the RNN\@. The set functions used also provide the inductive bias needed to incorporate the knowledge about the type of interaction: for example aggregation can be modeled by addition function, pooling by pointwise max function, etc.
    }
    \item {
        \textbf{Inroll}: The framework implicitly assumes that the rate of interaction between entities, i.e., flow of information on the graph, is equal to the rate at which we get to see observations at the nodes. This is not necessarily true in general. In particular if the latter rate is smaller as compared to the former, this can lead to poor performance. To correct for this, we introduce a parameter called \emph{inroll}. \emph{Inroll} is the number of times we unroll the network within the same time step, i.e., we repeat the input for that time step \emph{inroll} number of times, before feeding new input. Thus by choosing \emph{inroll} properly, we can ensure that the model has enough time to propagate the information to relevant nodes before having to process new inputs.
    }
\end{enumerate}

With these details in place, the framework allows us to use domain knowledge to appropriately share parameters and specify types of relationships between entities belonging to different classes. This allows gRNN models to learn the parameters to capture graph structured dynamics.

We note that this architecture can help us train models that have spatio-temporal generalization capability. Given a new scenario, if every entity in the scenario can be put into one of the equivalence classes, we can reuse parameters learned for that class. Further, if we construct summary graphs, in the same way that we did at training, we have a complete gRNN model which can be used to make predictions. Since the new graph also models the same kind of structures and interactions seen at training time, we can hope to generalize across time in the new scenario, thus achieving spatio-temporal generalization.

\section{Formal description} \label{Formal}

In this section we formally define scenarios and summary computation, and present the forward pass of a gRNN model in a given scenario.

\subsection{Scenario}
A scenario consists of a set of entities interacting in an environment. Formally it contains, the following elements:
\begin{enumerate}
    \item {
        $V$ is a set of nodes.
    }
    \item {
        $\mathcal{C}$ is a partition of $V$ with $C$ partitions or equivalence classes. $\mathcal{C} = \lbrace V_1, V_2, \ldots, V_C \rbrace$, with $\bigcup \limits_{i=1}^{C} V_i = V$ and $V_i \bigcap V_j = \phi, \forall \ i \neq j$.
    }
    \item {
        $T$ is the total time for which data is available in the current scenario. In our model, we only consider discrete time steps, and accordingly use $t$ to index over $[T]$. Generalizations to continuous time are not discussed here.
    }
    \item {
        We have one input data vector at each node at each time step. We assume that all nodes in the same equivalence class ($i$) have the same input dimension per time step ($p_i$). Thus, the input data is given by functions $X_i : V_i \times [T] \rightarrow \mathbb{R}^{p_i}$ for $i \in [C]$. Let $X = \lbrace X_1, X_2, \ldots, X_C \rbrace$.
    }
    \item {
        We also have one target vector per node, per time step. Again we assume that the output dimension is same for all nodes within an equivalence class. Let $q_i$ be the output dimension for nodes in the $i$-th equivalence class. Thus, the targets are given by functions $Y_i : V_i \times [T] \rightarrow \mathbb{R}^{q_i}$ for $i \in [C]$. Let $Y = \lbrace Y_1, Y_2, \ldots, Y_C \rbrace$.
    }
    \item {
        The relationships are defined using multiple graphs connecting the entities. We have one graph per relationship type. Let the edges in the $k$-th graph be denoted by $E_k \subseteq \lbrace (u, v) | u, v \in V \rbrace$. Let $E = \lbrace E_1, E_2, \cdots E_K \rbrace$ be the set of edges in the $K$ graphs.
    }
\end{enumerate}

For convenience we also define two more functions. $N(u, k)$ denotes the set of neighbors of node $u \in V$ in the $k$-th graph. Thus, $N(u, k) = \lbrace v \in V | (u, v) \in E_k \rbrace$. $g(u)$ denotes the equivalence class of the node $u$. Thus, $g(u) = i$ if $u \in V_i$.

\subsection{RNNs}
As described before, we put a RNN at each node, but we share the parameters across the equivalence class. Thus, we have $C$ RNNs. If $t$ is the current time step (ignoring \emph{inroll}), each of these RNNs take the current input ($x_t$) and the previous hidden state ($h_{t-1}$) and map it to the current output $y_t$ and the new state $h_t$. The $i$-th RNN has parameters $\theta_i$. We represent the RNNs as functions. Thus for $i \in [C]$, we have:
$$RNN_{\theta_i}(x_t, h_{t-1}) = (y_t, h_t)$$
We let all parameters be in the set $\Theta = \lbrace \theta_1, \theta_2, \ldots, \theta_C \rbrace $.

\subsection{Summary computation}
We assume that all RNNs have hidden state of the same dimension $d$. Summary for any node is an aggregation of hidden state of its neighbors from the previous time step. Since the number of neighbors is not known in a novel scenario, the summary aggregation function must have all finite subsets of $\mathbb{R}^d$ in its domain. We let $S$ be this set. $$S = \left\lbrace \left. \bigcup \limits_{i=1}^{n} \lbrace z_i \rbrace \right| n \in \mathbb{N}, z_i \in \mathbb{R}^d \right\rbrace \cup \lbrace \phi \rbrace$$

For each of the $K$ summary types, we use set functions defined on $S$ for summary computation. Let $m_k$ be the dimension of the $k$-th summary. Correspondingly, for $k \in [K]$, let $f_k : S \rightarrow \mathbb{R}^{m_k}$ be set functions with domain $S$. These will be applied to hidden states of the neighboring RNNs to compute summaries. We let $F = \lbrace f_1, f_2, \ldots, f_K \rbrace$.

\subsection{Prediction and objective}
Each RNN predicts one output vector at each time step at each node. Both target and prediction have dimension $q_i$ for nodes in the $i$-th equivalence class. We denote by $\widehat{y}_{u, t} \in \mathbb{R}^{q_i}$, the prediction by the model at node $u \in V$ at time $t \in [T]$. We use a loss function $\mathcal{L}_i : \mathbb{R}^{q_i} \times \mathbb{R}^{q_i} \rightarrow \mathbb{R}$ to measure the performance of our model for a single prediction. The total loss is an aggregation of all losses at all time steps. Thus, the loss can be written as
$$loss = \sum_{u \in V, \ t \in [T]} \mathcal{L}_{g(u)}(Y_{g(u)}(u, t), \ \widehat{y}_{u, t})$$
We let $\mathcal{L} = \lbrace \mathcal{L}_1, \mathcal{L}_2, \ldots, \mathcal{L}_C \rbrace$.

\subsection{Algorithm}

We are now ready to present forward pass algorithm for gRNN\@. For a given scenario, the forward pass is defined in Algorithm~\ref{gRNN_alg}.

We initialize all RNNs with zero state. Then at time step $t$, for every node $u$ and for every summary type $k$, hidden states from neighboring RNNs are aggregated and passed through summary function $f_k$ to get summary $s_{u, k, t}$. These are then concatenated with the input data to get the RNN input $x_{u, t}$. For every node, we do one step of RNN to get a new state vector $h_{u, t}$ and predictions $\widehat{y}_{u, t}$. This is repeated \emph{inroll} number of times for each time step. The predictions at the final \emph{inroll} step are compared to the targets to compute loss which is aggregated across nodes and time.

\begin{algorithm}
    \caption{Graphical RNN model: forward pass}\label{gRNN_alg}
    \begin{algorithmic}[1]
        \Require $V$, $\mathcal{C}$, $X$, $Y$, $E$, $F$, $\mathcal{L}$, $\Theta$, \emph{inroll} $= I$
        \State $\forall u \in V, h_{u, 0, I} = \vec{0} \in \mathbb{R}^d$
        \State $loss = 0$
        \For{$t = [1, 2, 3, \cdots T]$}
            \ForAll{$u \in V$}
                \State $h_{u, t, 0} = h_{u, t-1, I}$
            \EndFor
            \For{$i = [1, 2, 3, \cdots I]$}
                \ForAll{$u \in V$}
                    \ForAll{$k \in [K]$}
                        \State $s_{u, k, t, i} = f_k( \lbrace h_{v, t, i-1} | v \in N_{u, k} \rbrace )$
                    \EndFor
                    \State $s_{u, t, i} = concat \left( (s_{u, k, t, i})_{k=1}^{K} \right) $
                    \State $x_{u, t, i} = concat \left( X_{g(u)}(u, t), s_{u, t, i} \right)$
                \EndFor
                \ForAll{$u \in V$}
                    \State $( \widehat{y}_{u, t, i}, h_{u, t, i} ) = RNN_{\theta_{g(u)}} ( x_{u, t, i}, h_{u, t, i-1} )$
                \EndFor
            \EndFor
            \ForAll{$u \in V$}
                \State $loss = loss + \mathcal{L}_{g(u)}(Y_{g(u)}(u, t), \widehat{y}_{u, t, I})$
            \EndFor
        \EndFor
    \end{algorithmic}
\end{algorithm}

\section{Related Work} \label{RelWork}

A similar model has been proposed for learning on spatio-temporal graphs (\cite{jain2015structural}). Unlike our models however, the RNNs in their model cannot affect each other. In their setting, each node and each edge has an input feature vector at every time step. Since the RNNs on the edges do not take input from RNNs at nodes, the forward pass through their model is equivalent to running independent siloed RNNs, with one set of RNNs for each node. In contrast, we do not have any data input on edges. Additionally, in our case, summaries flow along each edge which are computed using hidden states of RNNs nearby and thus, all RNNs evolve together. Another assumption in their approach is that the same RNN on the edge provides input to RNNs on both sides. In our case, the edges can be directed and hence the interactions can be modeled differently. Finally we note that this model is a special case of our model with each edge replaced by a node and creating a disconnected graph with one connected component per node in the original graph.

GraphLSTMs have been proposed for semantic object parsing (\cite{liang2016semantic}). In this work, the pixels in an image are clustered into super-pixels and then a LSTM is applied at each super-pixel one by one. Unlike our case, the data being modeled is not sequential, but the summary computation has some similarities to our proposal. In their model, the LSTM at every node, also takes averaged hidden states of neighbors if that state has already been computed and ignored otherwise. Our proposal is differs from this work in terms of summary computations by incorporating (1) Equivalence classes of nodes and (2) Multiple summary types (3) Generic permutation invariant set transformations for summary computations.

A convolutional network that can work with arbitrary graphs has been proposed for learning molecular fingerprints (\cite{duvenaud2015convolutional}). Each node represents an atom and edges represent bonds. A feature vector is assigned to each possible atom type. For each convolutional layer, every atom adds activations from the previous layer of itself and its neighbors and then applies the same transformation on this aggregated input throughout the molecule at that layer. This model is a special case of our framework with all nodes in the same equivalence class, all edges in the summation summary, each RNN replaced by a convolution and inputs at all but the first step.

Liquid State Machines and Echo State networks are models of reservoir computing (\cite{jaeger2004harnessing, maass2002real}). Like gRNN, they also construct a connection graph with one RNN at each node. However, in these models, the nodes are randomly connected, and unlike our model, there is no way to inject domain knowledge into the system. One of the major criticisms of these models is that they are uninterpretable. In our model each node belongs to an equivalence class, and each summary flowing on any edge is a communication between nodes representing entities. Thus since the graph construction uses domain knowledge, same concepts may be useful to interpret what the model has learned. Additionally, these models assume fixed input dimension at every time step and cannot generalize if the input dimension increases. On the other hand, gRNNs share parameters within the class for any extra nodes that are introduced. Thus gRNNs organically handle varying input dimension for the network as a whole.

\section{Synthetic Data experiments} \label{Synth}

To test the abilities of models that use our framework, we generated some synthetic data. Every input dimension of the input is a sample from ARMA process \cite{whitle1951hypothesis} (as a function of time). For each of the models described below, many such processes are generated and fed into the model at various nodes.

The ARMA process was chosen since it is easy to characterize the optimal L2 loss for simple prediction problems. For example, if the input dimension is $3$ and we wish to predict the average of the three inputs at the next time step, the optimal L2 loss is $0.33$. Additionally, there is no other way to get this loss, because the targets depend on unobserved and independent random variables. Thus any other prediction other than the ARMA process function itself will have more variance and hence is not optimal. Thus checking whether the model achieves the optimal loss lets us test if has learned the data generating process.

\begin{figure}
    \centering
    \begin{subfigure}[b]{0.1\textwidth}
        \includegraphics[width=\textwidth]{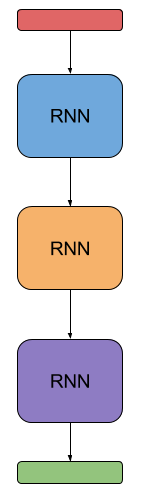}
        \caption{Hierarchical}\label{fig:Hierarchical}
    \end{subfigure}
    \hfill
    \begin{subfigure}[b]{0.3\textwidth}
        \includegraphics[width=\textwidth]{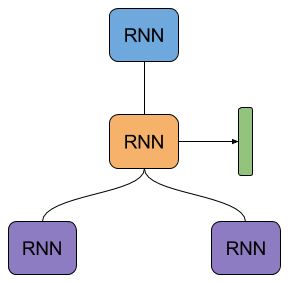}
        \caption{Tree 1}\label{fig:Tree1}
    \end{subfigure}
    \hfill
    \begin{subfigure}[b]{0.5\textwidth}
        \includegraphics[width=\textwidth]{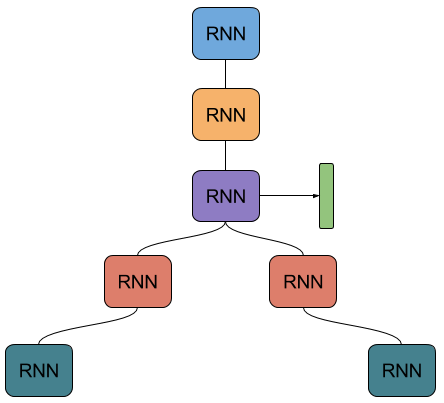}
        \caption{Tree 2}\label{fig:Tree2}
    \end{subfigure}
    \caption{Toy models. Undirected edges represent connections in both directions.}
    \label{fig:toy_models}
\end{figure}

For our experiments, we use one standard model and three gRNN models. These are described below and then some results on a simple prediction task using synthetic data are presented.
\begin{enumerate}
    \item {
        Basic: Regular RNN
    }
    \item {
        Hierarchical: A three level hierarchical model (see Fig.~\ref{fig:Hierarchical})
    }
    \item {
        Tree 1: (see Fig.~\ref{fig:Tree1}): We used two different summaries (one for connections going up and other for connections going down), one input dimension at each leaf and one at root, predict at center.
    }
    \item {
        Tree 2: (see Fig.~\ref{fig:Tree2}). Again, we used two different summaries (one for up and one for down), one input dimension at each leaf and one at root, predict at center.
    }
\end{enumerate}

The task is to predict at time $t$, the average of inputs fed into the model at the same time step, $t$. Even though the task is pretty simple, when inputs and outputs are at different nodes, it cannot be solved if we consider each node in isolation. Thus this task allows us to validate that gRNNs learn to properly route the information along the graph edges. Additionally, increasing the distance between the inputs and output locations also allows to test for \emph{inroll} functionality.

The results are presented in Table~\ref{tab:synth-result}. The gRNN models gets very close to the optimal loss values. Also, esults with higher \emph{inroll} suggest that the models have learned to effectively use this capability to get very close to the optimal losses.

\begin{table}
    \centering
    \begin{tabular}{| l | l | r | r |}
        \hline
        Model          & Experimental setup    & Optimal Loss  & Achieved Loss \\
        \hline
        Basic          & inroll=$1$            & $0$           & $\sim 0.001$ \\
        \hline
        Hierarchical   & inroll=$1$            & $0.33$        & $0.33 \pm 0.01$ \\
        \hline
        Hierarchical   & inroll=$2$            & $0$           & $\sim 0.0001$ \\
        \hline
        Tree 1         & inroll=$1$            & $0$           & $\sim 0.0007$ \\
        \hline
        Tree 2         & inroll=$1$            & $0.33$        & $0.33 \pm 0.02$ \\
        \hline
        Tree 2         & inroll=$2$            & $0$           & $\sim 0.008$ \\
        \hline
    \end{tabular}
    \caption{Results on synthetic dataset}
    \label{tab:synth-result}
\end{table}

\section{Weather prediction} \label{Weather}

Weather data is collected every day at weather stations across the globe. It is expected that these measurements are spatio-temporally correlated. Thus, knowledge of activity at nearby weather stations may help produce better forecasts. Accordingly, we can imagine an underlying graph connecting weather stations close to each other. This graph can provide useful information flow mechanism which our framework can exploit. We consider the task of predicting the average temperature on the next day at every weather station and optimize for mean squared error.\footnote{Temperature prediction was chosen because almost all weather stations collect this data and temperature is always available unlike some of the other meteorological measurements.}. On this task, we demonstrate the effectiveness of gRNN and show gains over strong baselines.

Google uses weather data for various products and hence the data was internally available in a structured form. Another data set for weather data is publicly available at \url{https://cloud.google.com/bigquery/public-data/noaa-gsod}. This public data is similar to the internal Google data on which we run our experiments.

\subsection{Setup}

\begin{figure}
    \centering
    \includegraphics[width=\textwidth]{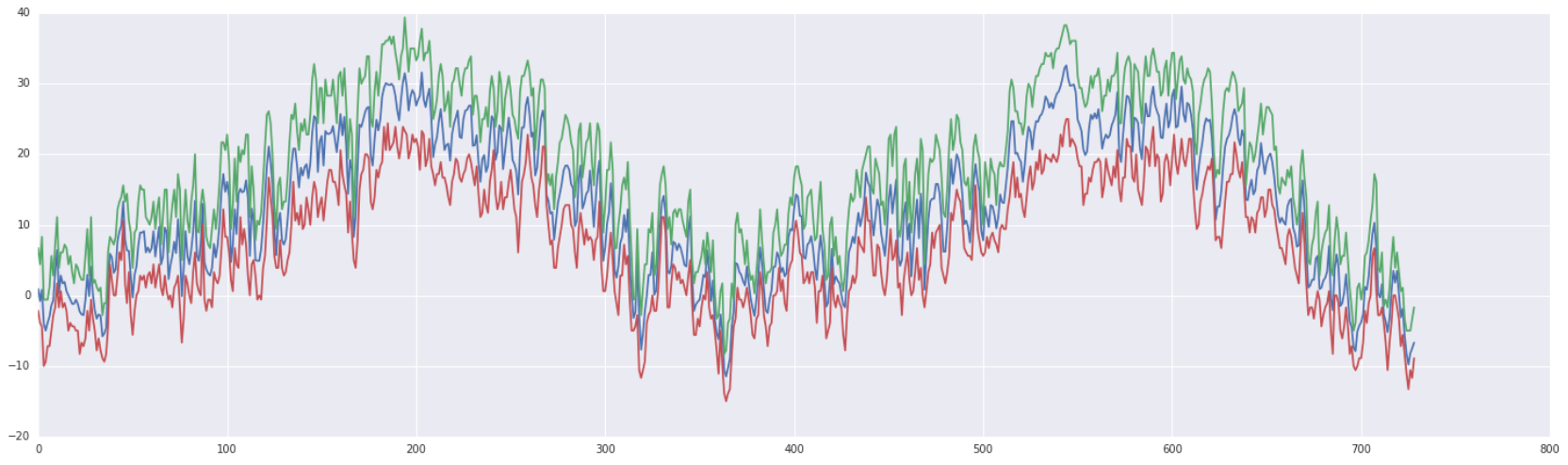}
    \caption{Sample input temperatures}
    \label{fig:temperature}
\end{figure}

Each day is one time step in our input data series. The weather measurements used are daily minimum, average and maximum temperature, i.e. three numbers per day at each weather station. We use data for the time period of two years (1st Jan 2014 to 31st Dec 2015) collected at more than 7000 stations across USA\footnote{Except Hawaii and Alaska, since they are too far from the mainland.}. A sample plot of the data can be seen in Fig~\ref{fig:temperature}.

As a pre-processing step, we removed the stations with a lot of missing data. For the remaining stations, we linearly interpolated to fill in the missing values. Further, stations with extreme data were then manually removed. This process was repeated till we got a reasonably clean dataset.

\begin{figure}
    \centering
    \includegraphics[width=\textwidth]{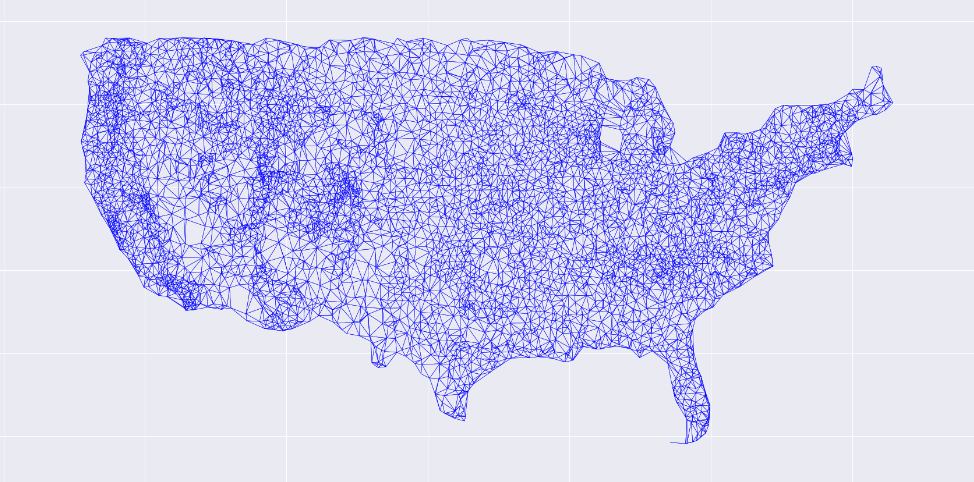}
    \caption{Triangulation graph}
    \label{fig:tri_graph}
\end{figure}

For each weather station, we also have its physical location on the map as given by the latitude and longitudes. We use these values to put every weather station on a 2D map. Then we construct a Delaunay Triangulation using weather stations as nodes. To remove long edges which are artifacts of the triangulation process, we then thresholded the edge distances such that more than 95\% edges were preserved. This gives us the final graph which was used for the experiments. (See Fig~\ref{fig:tri_graph}).

\begin{figure}
    \centering
    \includegraphics[width=\textwidth]{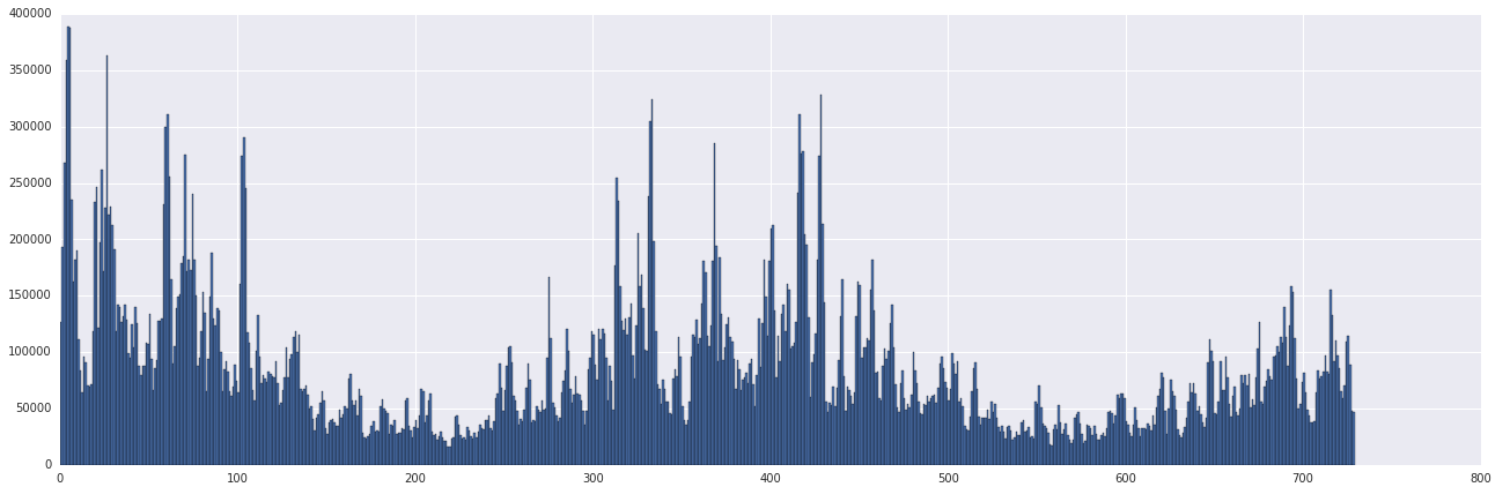}
    \caption{Steady state prediction error}
    \label{fig:ss_pred_err}
\end{figure}

\subsection{Experiments and Results}
As mentioned above, our task it to predict the average temperature on the next day. We tried three kinds of models:
\begin{enumerate} \itemsep0em
    \item Simple models: They make point predictions (based on current input), and cannot leverage historical patterns
    \item RNN models: These models can take historical trends into account and leverage function approximation capabilities of neural networks. However they do not leverage information from nearby stations.
    \item gRNN models: In addition to using history and powerful function approximation, these models make use of the context around them as given by the triangulation graph.
\end{enumerate}

As a baseline, we use a steady state prediction model. This model predicts the average temperature at next time step to be the same as the current average temperature. This model achieves mean squared L2 error of $11.87$ on year 2014 and $9.79$ on year 2015. There is a significant gap in these numbers which points to a skew in the dataset\footnote{Some of the variations are seasonal (Fig. \ref{fig:ss_pred_err} )}.

We train on data for the year 2014 and test on 2015. While testing, the RNN and gRNN models are still fed the 2014 data, but they start predicting only on 2015. We used squared L2 loss for training as well as testing. All RNN and gRNN models were trained with Truncated Backpropagation Through Time (\cite{williams1990efficient}) due to long length of the sequences (730 days). We tried LSTM with tanh activations and iRNN (\cite{le2015simple}) with ReLU activations. We also tried LSTMs with ReLU, but they were highly unstable and quickly ran into NaNs. For gRNN, we use the same equivalence class for all weather stations, since we did not know any further classification. We also compared summation and averaging summaries within gRNN models.

The results are summarized in Table \ref{tab:weather-results}. We report the errors for all models as a percentage of the corresponding error achieved by steady state prediction model. As expected, the linear model is better than the steady state model. LSTM with tanh performs better than the linear model and iRNNs with ReLU activations outperform LSTMs. Within gRNN models, iRNN with ReLU activation using the triangulation graph and summation summary performs better than not using the graph. Finally, using average summary with iRNN ReLU performs the best on both training as well as test set by quite a margin. Average summary might be better suited for this task since the density of weather stations is not be uniform and averaging normalizes for this variation. Overall, we see substantial gains going from simple models, to RNNs to gRNNs.

\begin{table}
    \centering
    \begin{tabular}{| l | l | l | r | r |}
        \hline
        Model type              & Details                         & Train L2  & Test L2 \\
        \hline
        \multirow{2}{*}{Simple} & Steady state prediction         & $100.00$   & $100.00$  \\\cline{2-4}
                                & Linear                          & $ 97.22$   & $ 97.34$  \\
        \hline
        \multirow{2}{*}{RNN}    & LSTM, tanh                      & $ 90.48$   & $ 92.75$  \\\cline{2-4}
                                & iRNN, ReLU                      & $ 87.62$   & $ 91.01$  \\
        \hline
        \multirow{2}{*}{gRNN}   & iRNN, ReLU, summation summary   & $ 84.41$   & $ 89.38$  \\\cline{2-4}
                                & iRNN, ReLU, average summary     & $ 81.80$   & $ 87.33$  \\
        \hline
    \end{tabular}
    \caption{Results on weather prediction task (\% of steady state baseline).}
    \label{tab:weather-results}
\end{table}

\section{Discussion and Future Work} \label{DiscussFuture}

As shown by the empirical results (table \ref{tab:weather-results}), we think that Graphical RNN models provide a useful framework for learning from graph structured processes. Below, we present a few ideas that can be used to refine the gRNN approach.

\subsection{Graph Sampling for Stochastic Gradient}
Training gRNNs can be slow because they need to update all the states at every time step. Here, we present an idea to speed them up by using graph sampling. The main idea is to compute a random ``overview'' graph on the fly to generate different overviews of the whole graph and treat each of them as an example. An overview graph should ideally (1) have small number of nodes and edges (2) the expectation of the minibatch stochastic gradient computed on this random graph should equal to the true gradient, where the expectation is both with respect to the minibatch sampling as well as the random sampling from the space of all overview graphs. Note that the second condition is crucial for stochastic gradient descent algorithms to converge.

If we are able to do this, this will reduce the computation per time step and can potentially scale to very large gRNNs, such as social network graphs. To use this method in practice however, we need an efficient randomized overview graph sampling algorithm which satisfies the necessary properties. Below we describe two concrete examples where it is easy to imagine such a randomized algorithm.
\begin{enumerate}
    \item{
        For weather data: Sample the nodes randomly and construct a new triangulation on them. Since the nodes will be sampled uniformly, we can expect that the triangulation thus produced will be a good overview.
    }
    \item{
        For trees structured gRNN\@: Sample at the first level below the root. Then sample children of selected nodes, and so on.
    }
\end{enumerate}
These examples are natural because of domain knowledge or known graph structure. But it is not clear what should be done for general graphs, and the design of such an algorithm for generic gRNNs is left as an open question.

Note that we may need to do some changes while operating on the overview graph. For example, if we are using summation as the permutation invariant transform, we need to scale the summaries on subsampled graph accordingly since the new expected degree of the nodes can be different than before. For the overview graph, the \emph{inroll} can decrease as we subsample more since there are fewer hops between nodes. This is again problem and sampling dependent. For example, in tree structured gRNNs, \emph{inroll} should remain the same, but for weather prediction task, it should decrease.

\subsection{gRNN with attention}

For each node, before combining adjacent hidden states into summaries we can reweigh them by score that indicates how useful the hidden state along the incoming edge is. Like an attention model, these weights can be learned through a scoring function that takes as input both hidden states. Note that more generally, this means that in Algorithm~\ref{gRNN_alg}, line number 9 should now become
$$s_{u, k, t, i} = f_k( \lbrace h_{v, t, i-1} | v \in N_{u, k} \rbrace | h_{u, t, i-1} )$$ denoting that the function $f_k$ can now use the previous hidden state of the node $u$, but should still satisfy the constraints outlined in Section~\ref{summary-desc}. These functions can now additionally have learnable parameters as well, which we can train jointly with other parameters.

We note that gRNN with attention is more powerful than the original formulation since the former is the same as latter when the functions $f_k$ ignore $h_{u, t, i-1}$.

\section{Conclusion}

In this paper, we proposed Graphical RNN models (gRNN), a framework to design model architectures capable of spatio-temporal generalization. We described several mechanisms afforded by our framework to incorporate domain knowledge into the model architecture design. We validate our approach on several toy tasks meant to illustrate the features of our approach. We showed the effectiveness of our approach on a weather prediction task by empirically demonstrating improvements over other approaches. Finally, we discussed potential approaches to make gRNNs faster and more powerful.

\bibliography{grnn_jmlr}
\end{document}